\documentclass[11pt]{article}

\usepackage[a4paper,top=1in,bottom=1in,left=1.1in,right=1.1in]{geometry}
\usepackage{setspace}
\usepackage[T1]{fontenc}
\usepackage[utf8]{inputenc}
\usepackage{textcomp}
\usepackage{lmodern}
\usepackage{tgheros}

\usepackage{microtype}
\usepackage{dirtytalk}
\usepackage{tikz}
\usetikzlibrary{arrows.meta}

\usepackage{amsmath, amssymb}

\usepackage{xcolor}
\definecolor{accent}{HTML}{1F4E79}
\definecolor{rulegrey}{HTML}{B7B7B7}
\definecolor{captionnotation}{HTML}{555555}

\usepackage{graphicx}
\graphicspath{{figs/}}
\usepackage[font=small,labelfont={bf,color=accent},labelsep=period,format=plain]{caption}
\usepackage{subcaption}

\usepackage{booktabs}
\usepackage{array}
\newcolumntype{L}[1]{>{\raggedright\arraybackslash}p{#1}}

\usepackage[round,authoryear]{natbib}
\usepackage[hidelinks,colorlinks=true,
            linkcolor=accent,citecolor=accent,urlcolor=accent]{hyperref}

\usepackage{titlesec}
\titleformat{\section}[hang]
  {\Large\bfseries\color{accent}}{}{0pt}
  {}
  [\vspace{-3pt}{\color{rulegrey}\titlerule[0.6pt]}]
\titlespacing*{\section}{0pt}{18pt}{8pt}
\titleformat{\subsection}[hang]
  {\large\bfseries\color{accent}}{}{0pt}{}
\titlespacing*{\subsection}{0pt}{14pt}{4pt}
\titleformat{\subsubsection}[runin]
  {\normalsize\bfseries\color{accent}}{}{0pt}{}[.]
\titlespacing*{\subsubsection}{0pt}{6pt}{6pt}

\newcommand{\alphaval}{\ensuremath{\alpha}}
\newcommand{\Qstar}{\ensuremath{Q^{\!*}}}
\newcommand{\SOnn}{\ensuremath{\mathrm{SO}(n)}}

\usepackage[normalem]{ulem}

\begin{document}

\pagestyle{plain}\thispagestyle{empty}
\begin{center}
  {\color{accent}\fontsize{18}{22}\selectfont\bfseries
    Partially canonical axes \\ in large language model representations\par}
  \vspace{0.8em}
  {\normalsize
    Andrea Gregor de Varda\textsuperscript{1,2,*},\quad
    Ferdinando D'Elia\textsuperscript{3},\quad \\
    Greta Tuckute\textsuperscript{4},\quad
    Evelina Fedorenko\textsuperscript{1,2,\dag},\quad
    Meenakshi Khosla\textsuperscript{5,\dag}\par}
  \vspace{0.4em}
  {\footnotesize\itshape\color{captionnotation}
    \textsuperscript{1}McGovern Institute for Brain Research, Massachusetts Institute of Technology\\[1pt]
    \textsuperscript{2}Department of Brain and Cognitive Sciences, Massachusetts Institute of Technology\\[1pt]
    \textsuperscript{3}Center for Language Technology, University of Copenhagen\\
    \textsuperscript{4}Kempner Institute, Harvard University\\[2pt]
    \textsuperscript{5}Department of Cognitive Science, University of California San Diego\\[2pt]
    \textsuperscript{*}\textnormal{Correspondence: \href{mailto:devar_ag@mit.edu}{devar\_ag@mit.edu}} \hspace{.6cm} \textnormal {\textit{\textsuperscript{\dag}These authors jointly supervised this work.}}}
    
\end{center}
\vspace{0.8em}

\begingroup
{\small 
\begin{quote}

\noindent Large language models (LLMs) develop rich internal representations of language, but whether their individual coordinate axes carry meaning, or whether meaning is encoded only in directions in representation space, is still an open question. A common assumption is that individual axes in LLMs are arbitrary. We here challenge this view: across 12 LLMs from six model families, the native axes are reliably more aligned with other models' representations, with human brain responses to language, and with interpretable linguistic features than randomly rotated axes that preserve the same geometry. The native axes nonetheless fall well short of the best achievable alignment, so LLM coordinates are not fully arbitrary but also not fully canonical: they are partially canonical. A theory of how LLMs represent language must explain both the population geometry they learn and why individual units acquire patterns of response profiles that recur across systems.
\end{quote}}
\vspace{2pt}
\endgroup

\section{Introduction}

Large language models (LLMs) now master linguistic abilities and produce text that human judges struggle to distinguish from human writing \citep{jones2025large}. These abilities are supported by hidden representations that encode a hierarchy of linguistic features, from sublexical information, to syntactic structure, to compositional meaning \citep{tenney2019bert, jawahar2019what}. Such representations can be seen as points in a high-dimensional space, and the model's \textit{native basis} describes those points with one axis per unit (one \say{neuron}). Two lines of work have tried to explain \emph{how} this information is encoded in such representations. 

The first approach focuses on individual neurons, and aims to find specialized units in the models' representation spaces that selectively respond to certain linguistic features. This line of work predates transformer LLMs. Within this program, several studies have found that individual neurons in LLMs (or small sets of them) end up specializing for particular aspects of language processing, from shallow form-level features such as token position or punctuation \citep{bauidentifying, gurneefinding, voita2024neurons} to morphological \citep{bauidentifying, gurneefinding, antvergpitfalls, han2026modular} and syntactic features such as number agreement \citep{lakretz2019emergence, mueller2022causal, varda2023data, kryvosheieva2025syntactic, han2026modular}, particular kinds of meaning, such as affective content \citep{radford2017sentiment, devarda2024emergence}, specific natural or programming languages \citep{gurneefinding}, or even factual associations \citep{meng2022locating, dai2022knowledge, gurneefinding} and gender bias \citep{vig2020investigating}. 
Throughout this approach, the causal contribution of units to particular computations is tested with targeted ablations that selectively disrupt the features encoded by those units \citep{lakretz2019emergence, kryvosheieva2025syntactic}. 

The second approach focuses on populations, treating the hidden state as a whole and recovering interpretable content from directions in activation space. Here, a single meaningful feature can be spread across many---and possibly all---units in the original coordinate system: the relevant directions need not coincide with the original coordinate axes. (Note that a distributed feature \textit{could} in principle still be confined to multiple native units, spanning a multidimensional subspace aligned with the native basis \citep{mueller2025quest}; here we focus on the case where it is not.) This \say{directional} view has its origin in word embeddings, where simple arithmetic over vectors recovers analogies such as ``queen = king -- man + woman'' \citep{mikolov2013efficient}, and richer knowledge about objects (such as their size or level of danger) can be read out by projecting word vectors onto interpretable axes \citep{grand2022semantic}. The same approach has been applied to contextualized representations. For instance, dependency trees can be recovered from a learned linear projection of BERT's hidden states, showing that syntactic structures are realized in a subspace that spans many units in the native basis \citep{hewitt2019structural}. Similar approaches have been applied to semantics: for instance, the truth or falsity of a factual statement \citep{marksgeometry} or its plausibility \citep{bertolazzi2025language} are encoded as linear directions in the residual stream. Under this second approach, interventions act on directions instead of coordinates, for instance steering behavior by adding or subtracting a feature vector from the hidden state instead of manipulating any single unit \citep{turner2023steering, rimsky2024steering}. 

Support for the directional view comes from polysemanticity, the observation that individual units respond to multiple, apparently unrelated features (\citealp{bricken2023towards}, cf. \citealp{cammarata2026clean}). This pattern has been hypothesized to arise from \emph{superposition}, whereby models can encode more features than they have dimensions by representing features as non-orthogonal directions distributed across units \citep{arora2018linear, elhage2022toy, mueller2025quest}. The directional view has been formalized in the Linear Representation Hypothesis for modern LLMs \citep{park2024linear}, whereby high-level attributes of the input are encoded linearly as directions in the model's representation space \citep[see][for an account of why this should be the case]{klindt2026unifying}. This idea underlies the most historically successful programs in mechanistic interpretability, including probing classifiers \citep{belinkov2022probing} and sparse autoencoders \citep{cunningham2023sparse, templeton2024scaling}. However, recent mechanistic interpretability work has argued that this picture should not be taken for granted in transformers: because some operations act coordinate-wise, the residual stream may possess a privileged basis, and optimization applied to coordinates (e.g., the Adam optimizer; \citealp{adam}) may further encourage features to align with the native basis \citep{elhage2023privilegedbasis, adamprivilegedbasis}.

A similar debate on whether information is encoded in single units or in the population geometry has been going on for decades in neuroscience \citep{eichenbaum2018barlow}. One tradition holds that individual neurons are tuned to specific features, as with cells tuned to orientations in the primary visual cortex \citep{hubel1968receptive}. This kind of selective tuning is found in higher-level brain regions too \citep{tsao2003faces}, where many neurons or voxels respond consistently to the same feature across individuals \citep{khosla2024privileged}, and some neurons in the human medial temporal lobe fire for one particular person or object across different representation formats \citep{quiroga2005invariant}. According to an alternative view, what matters are \textit{patterns} of activation across a whole population and their geometry, with each neuron contributing to many (and possibly all) features. Recordings in frontal lobes of macaques, for example, have shown that many neurons show \emph{mixed selectivity} (the biological counterpart of polysemanticity): they respond to several task variables at once, and their tuning shifts with context and task demands \citep{rigotti2013importance, mante2013context}. So here too, whether the individual units carry meaning on their own or only the population geometry does is an open question that depends on the region and the task \citep[see][for a review]{eichenbaum2018barlow}.

Besides the conceptual debate about the nature of representations in AI systems, this distinction also has practical implications because it constrains how model representations can legitimately be studied and which interpretability tools are licensed. If coordinates are fully arbitrary, then analyses of individual units and the family of methods that read information off single neurons recover nothing that a random direction would not, and only direction- or subspace-level methods are meaningful; if instead coordinates are privileged, neuron-level analyses are justified, and, conversely, purely rotation-invariant analyses ignore meaningful structure in the models' representations. Knowing whether LLMs' axes are meaningful can also inform how representations are compared at all: the standard metrics for model-to-model and model-to-brain comparisons (such as representational similarity analysis; \citealp{kriegeskorte2008representational} based on Euclidean distances) are by construction invariant to rotations of the basis and thus blind to a component of representational agreement.

One may be tempted to conclude that the existence of polysemanticity in LLM units already answers this question: if each unit responds to many unrelated inputs, single units, and, by extension, axes, could not be meaningful. However, this conclusion does not follow: although a unit may code for several things at once, it could still show a preference for particular information in a way that goes beyond what a random linear combination of all units could encode. To our knowledge, whether LLM coordinate axes are fully arbitrary, fully canonical, or somewhere in-between has not been tested directly. \citet{khosla2024privileged} recently found that the native bases of vision and audition deep networks and of sensory cortices are privileged: they are reliably more aligned across systems than randomly rotated versions are. However, the analogous question for neural network language models, where the linear representation hypothesis has been most influential, has remained open.

\begin{figure}[t]
  \centering
  \resizebox{\textwidth}{!}{\input{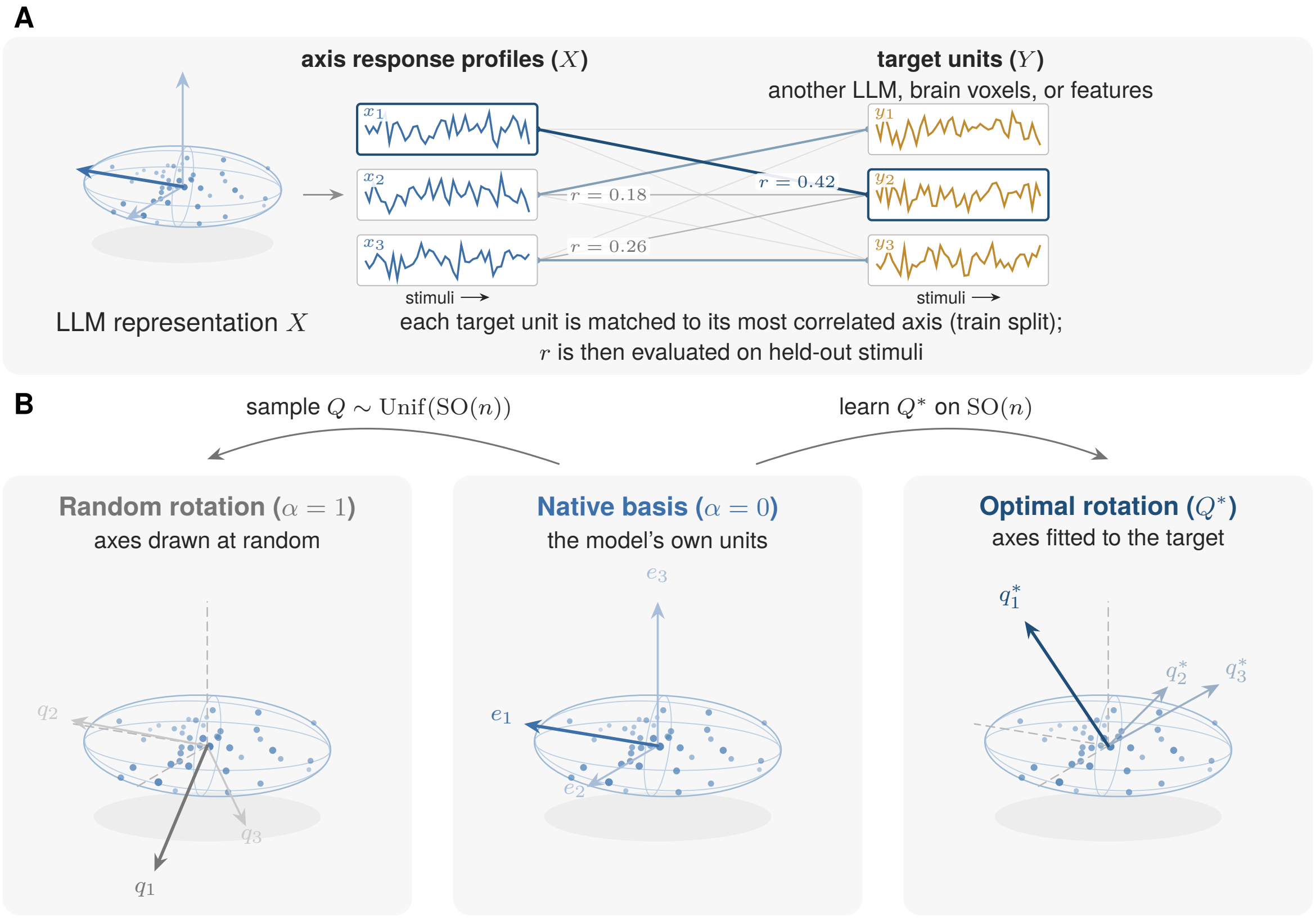}}
  \caption{\textbf{Axis-by-axis alignment under rotation of the coordinates.} \textbf{A.} Schematic of the unit matching procedure. Each coordinate axis of an LLM hidden representation $X$ (one axis per unit) generates a response profile across the input stimuli. Each unit of a target representation $Y$ (a unit of another LLM, a voxel in the brain, or an interpretable linguistic feature) is matched to its most correlated axis in $X$ on a training split; the alignment score is the correlation $r$ between matched pairs on held-out stimuli, averaged over target units in $Y$. \textbf{B.} The stimuli in $X$ are represented as a point cloud in three coordinate frames (with all pairwise distances fully preserved). \textbf{Left:} a random orthogonal rotation ($\alphaval = 1$) moves away from the native axes but keeps the geometry intact. \textbf{Middle:} the native basis ($\alphaval = 0$), where each axis corresponds to one model unit. \textbf{Right:} the optimal rotation $\Qstar$, learned by gradient descent on \SOnn{} to maximize the pairwise alignment with the target. If the native axes were arbitrary, the matching in \textbf{A} would give identical alignment scores in the left and middle panel; if they were fully canonical, it would give identical scores in the middle and right panels.}
  \label{fig:logic}
\end{figure}

The single-unit tradition described above combines two claims that can, and should, be distinguished. In the first, an axis is privileged if it recurs across independent instantiations of a system---across random initializations, across architectures, or across substrates as different as an artificial network and a brain. This is a claim about axis \textit{convergence} and can be assessed without any external notion of meaning. In the second, an axis is privileged if it aligns with an interpretable feature: a property such as sentiment, number agreement, or a perceptual dimension. This is a claim about \textit{content}, and it requires an external target to align against. Although the two senses are independent, there is one procedure that can be applied to test for both (Fig.~\ref{fig:logic}). Rotating an LLM's representation with an orthogonal transformation reorganizes the coordinate frame while preserving every pairwise distance. If alignment to a target representation, be it another representational system or an interpretable feature, is unaffected by this rotation, the native axes are arbitrary; the particular axes matter and not only the distances between points.

By applying such an axis-sensitive analysis to LLM representations, we find that the native coordinate frame of an LLM is \emph{partially} canonical. Across many trained models and across several alignment targets that include other LLMs, human fMRI responses in the language network \citep{pereira2018toward}, and interpretable linguistic features, the native basis of LLM representations is consistently more aligned with those targets than a random, geometry-preserving variant obtained with rotation. So individual axes do carry information that a rotation-invariant analysis cannot recover. At the same time, the native basis falls short of the best alignment achievable by any rotation of its axes: LLM axes are not fully canonical either.

\section{Results}

\begin{figure}[t]
  \centering
  \includegraphics[width=\textwidth]{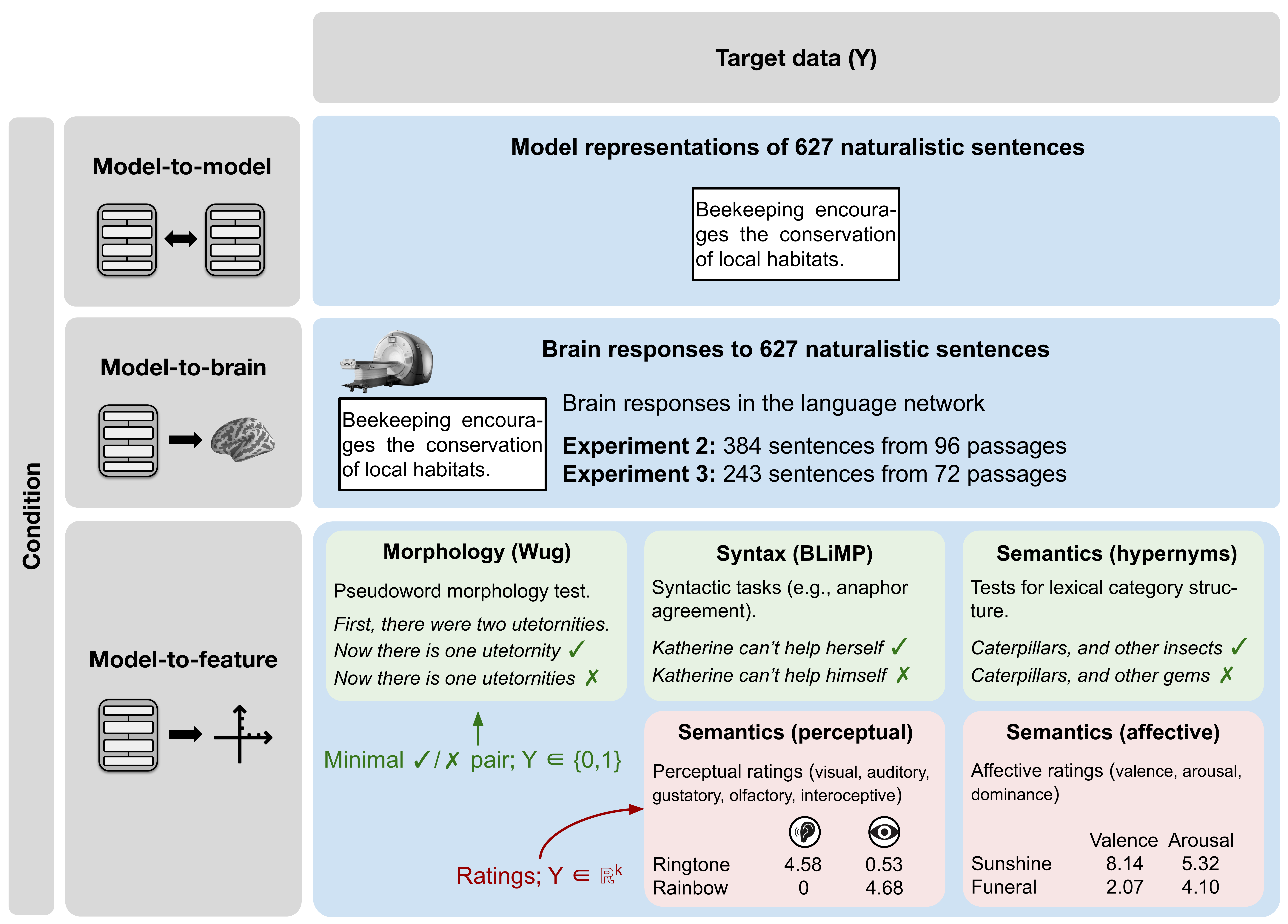}
  \caption{\textbf{Three settings for probing the LLM representation basis.} In each setting an LLM's hidden representation ($X$) is aligned to a target representation ($Y$) via axis-by-axis matching.  For each basis of $X$ (native or rotated), we match each target dimension to the single axis of $X$ with which it has the highest absolute correlation on training stimuli and evaluate that match on held-out stimuli. \textbf{Top:} \emph{Model-to-model}, where $Y$ is the hidden representation of another LLM elicited by the same set of 627 naturalistic sentences (the 384 sentences of Pereira Experiment 2 combined with the 243 sentences of Pereira Experiment 3). \textbf{Middle:} \emph{Model-to-brain}, where $Y$ is human fMRI responses recorded in the language network while participants read the same 627 sentences across two experiments. \textbf{Bottom:} \emph{Model-to-feature}, where $Y$ is an interpretable linguistic feature, either a binary well-formedness label from a minimal-pair task ($Y \in \{0, 1\}$; morphology, syntax, hypernymy) or a multi-dimensional rating vector ($Y \in \mathbb{R}^k$; perceptual and affective ratings).}
  \label{fig:overview}
\end{figure}

\subsection{Framework and overview}

We analyze 12 language models from six model families: GPT-2 in four sizes (124M, 355M, 774M, 1.5B; \citealp{radford2019language}); LLaMA-3.2 in two sizes (1B and 3B; \citealp{grattafiori2024llama}); Gemma-3 (1B; \citealp{gemmateam2025gemma3technicalreport}); Mamba in three sizes (130M, 370M, 790M; a state-space instead of a transformer architecture; \citealp{gu2023mamba}); Qwen-3 (0.6B; \citealp{yang2025qwen3}); and Pythia (160M, including a second 160M model trained from a different random seed; \citealp{biderman2023pythia, van2025polypythias}). We tested whether LLM representations have privileged axes using three complementary comparisons spanning model-to-model, model-to-brain, and model-to-feature alignment (Fig.~\ref{fig:overview}). First, we compared the activations of two different LLMs to each other, asking whether independently trained models agree on the same coordinate axes, using 627 naturalistic sentences (\emph{model-to-model alignment}; Fig.~\ref{fig:overview}, top row). Second, we compared LLM activations against human fMRI responses to the same 627 naturalistic sentences \citep[data from][]{pereira2018toward} recorded from the language network \citep{fedorenko2024language} while participants read natural sentences (\emph{model-to-brain alignment}; Fig.~\ref{fig:overview}, middle row). Third, we compared LLM activations against interpretable linguistic features (\emph{model-to-feature alignment}; Fig.~\ref{fig:overview}, bottom row). For the latter analysis, we consider features that span morphology (\say{Wug} task, using data from \citealp{hanna2026formal}), syntax (BLiMP, using 12 sub-tasks; \citealp{warstadt2020blimp}), and lexical semantics (hypernymy minimal pairs adapted from \citealp{hanna2021analyzing}; Lancaster sensorimotor norms, \citealp{lynott2020lancaster}; and ratings of valence, arousal, and dominance, \citealp{warriner2013norms}).

We quantify alignment using the axis-by-axis matching procedure illustrated in Fig.~\ref{fig:logic}A, introduced by \citet{khosla2024privileged}. For two representations---$X$ (LLM activations) and $Y$ (activations from other LLMs, brain responses, and linguistic features)---over the same set of $n$ stimuli, we find for each unit (column) in $Y$ its single best-correlated unit in $X$ on a training split (matching on the absolute correlation), then report the Pearson $r$ between the matched pairs on a held-out test split. To ask whether alignment depends on the specific axes of $X$, we smoothly rotate its coordinates, preserving its representational geometry (Fig.~\ref{fig:logic}B). Specifically, we apply a smooth rotation $Q^{\alpha}$ to $X$, where $\alphaval = 0$ corresponds to no rotation and $\alphaval = 1$ corresponds to a fully random orthogonal rotation drawn uniformly from the special orthogonal group \SOnn{} (i.e., the set of rotations excluding reflections). We additionally compute an \emph{optimal axis alignment} score (Fig.~\ref{fig:logic}B, right): an orthogonal $\Qstar$ is learned by gradient descent on the \SOnn{} manifold to maximize the matching score on held-out validation data, giving an estimate of the best axis-by-axis alignment achievable by rotating the basis of $X$. Throughout, alignment is averaged across layers (excluding the embedding layer; see SI~\ref{sec:si-layers} for validation in individual layers) and reported as Pearson $r$, averaged over the dimensions of $Y$. 

If LLM coordinates were arbitrary, alignment would be invariant to $\alphaval$ (the amount of axis rotation). If individual axes were perfectly aligned with our target dimensions in $Y$, alignment at $\alphaval = 0$ would match the optimal axis alignment. Instead, we found that the native basis sits between a random rotation and the optimum in essentially every comparison we performed -- supporting a \emph{partially canonical} characterization of LLM coordinates.

\subsection{Model-to-model alignment}

\begin{figure}[t]
  \centering
  \includegraphics[width=\textwidth]{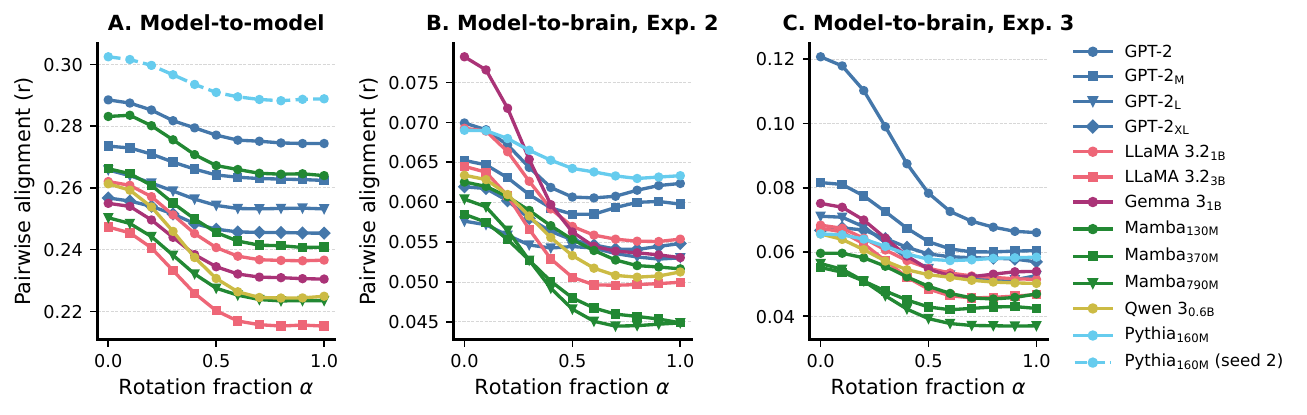}
  \caption{\textbf{Model-to-model and model-to-brain alignment as a function of rotation.}
  \textbf{A.} Pairwise alignment between Pythia-160M and every other model in the set, as a function of the rotation fraction $\alpha$. The light-blue dashed line is the comparison with an identical model trained with a different seed. \textbf{B, C.} Pairwise alignment of each LLM with human fMRI responses in the language network, separately for Experiment 2 (B) and Experiment 3 (C) of \citet{pereira2018toward}. Each curve is the mean over layers (and participants); alignment decreases with $\alpha$ in most models and in every panel on average.}
  \label{fig:1}
\end{figure}

The most theory-agnostic version of our question is whether representational axes are aligned across different LLMs, given that it does not involve any external reference to evaluate the tuning of the axes (Fig.~\ref{fig:overview}, top row). We use Pythia-160M as the reference model and align it pairwise with all the other models. For every pair the matching score is computed in both directions (rotating the basis of one model while keeping the other fixed, then vice-versa) and the two values are averaged. For this comparison (and for the training dynamics analysis below), we averaged the alignment scores over four layers at four relative depths (25, 50, 75 and 100\%). 

Across every model pair, alignment declines as $\alphaval$ increases (Fig.~\ref{fig:1}A): when each unit of a target LLM is read out from the most correlated unit in the source LLM, native axes recover more of that unit's variance than randomly rotated axes. Averaging across pairs, alignment in the native basis is 0.27 and it drops to 0.25 at a full random rotation (native -- random = 0.021, 95\% CI [0.016, 0.027]; positive for 12/12 pairs, Wilcoxon \textit{p} = 5 $\times$ 10\textsuperscript{-4}); alignment declines with $\alpha$ in every pair (mean Kendall $\tau = -0.93$; Stouffer \textit{Z} = 16.5, \textit{p} < 10\textsuperscript{-60}). The gap is numerically small but extremely consistent across models. The uppermost curve in Fig.~\ref{fig:1}A (light blue) is the comparison with a model with the same architecture and size (Pythia-160M), but pre-trained with a different seed; the two models are noticeably better aligned in their native basis than any other pair of models (0.30 native), but the drop from native to fully-rotated basis (which diagnoses whether the native basis is privileged) is comparable in magnitude to that of the other pairs. Thus, the pattern is not specific to pairs of models with the same architecture and emerges with similar magnitude across model families.

\subsection{Model-to-brain alignment}

We next ask whether the same partially canonical pattern holds when we align LLM activations to responses in the human brain, which until recently was the only system capable of processing language (Fig.~\ref{fig:overview}, middle row). We use fMRI data from \citet{pereira2018toward}, recorded in the language network -- a set of left-lateralized frontal and temporal regions defined in each participant individually using a localizer task \citep{fedorenko2010new}. We restricted our analyses to the
top-10\% language-selective voxels per participant, averaged each voxel's responses over the repeated presentations of each sentence, and kept the 20\% of these voxels with the highest test-retest reliability (Methods; see \ref{sec:si-fmri} for similar patterns using all language-responsive voxels). We consider data from two experiments where sentences (384 sentences from 96 passages in Experiment 2, 243 sentences from 72 passages in Experiment 3) were presented one at a time. We leave entire passages out in cross-validation to avoid leakage between training and test sentences \citep{hadidi2026spurious}. For each experiment, the LLM representations ($X$) are rotated by $Q^{\alpha}$ while the voxel responses ($Y$) are held fixed---in other words, we ask how well each individual voxel can be predicted from its single best-matched (rotated) LLM unit.

Across both experiments, alignment with brain responses decreases as $\alphaval$ increases (Fig.~\ref{fig:1}B, C; Stouffer \textit{Z} = 13.5 and 14.2, both \textit{p} < 10\textsuperscript{-41}). Relative to the native basis, a full random rotation reduces mean alignment by approximately 17\% in Experiment 2 (from $r = 0.065$ to $0.054$) and 27\% in Experiment 3 (from $r = 0.071$ to $0.052$). Although the absolute changes in $r$ are modest, all 12 models show lower alignment after rotation in both experiments (Wilcoxon signed-rank $p = 5 \times 10^{-4}$ in each).

\subsection{Model-to-feature alignment}

\begin{figure}[t]
  \centering
  \includegraphics[width=\textwidth]{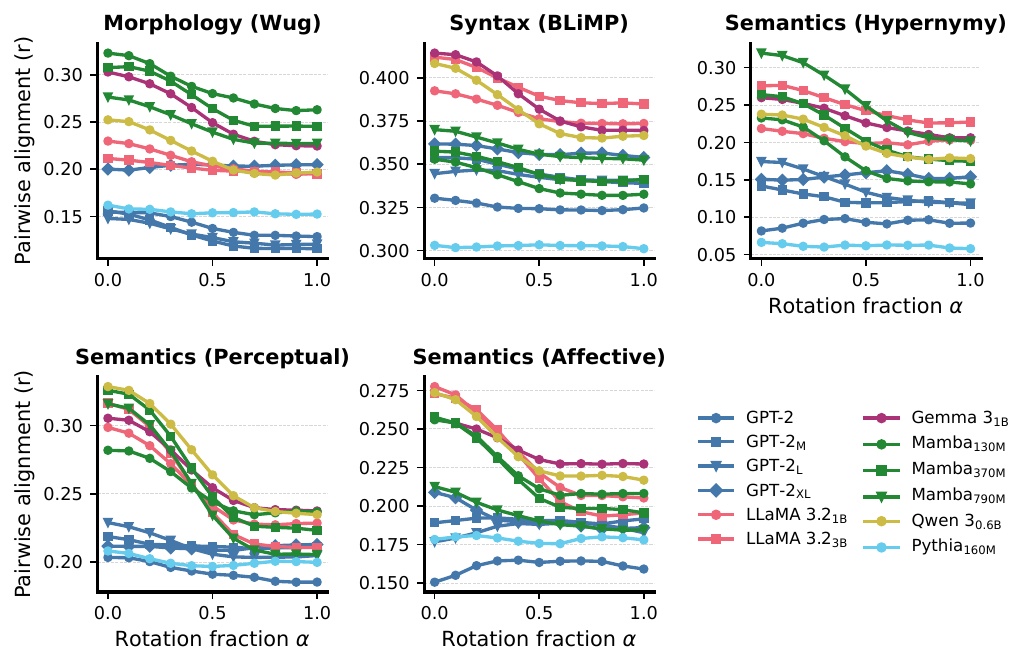}
  \caption{\textbf{Model-to-feature alignment as a function of rotation.} Pairwise alignment of each LLM against five interpretable feature targets that span morphology (Wug), syntax (BLiMP, averaged across 12 sub-tasks) and lexical semantics (hypernymy minimal pairs; perceptual sensorimotor norms; affective valence/arousal/dominance norms). Each curve is the mean over layers; the same decline with $\alpha$ appears in all panels.}
  \label{fig:2}
\end{figure}

Beyond aligning to other representational systems, we wanted to know whether individual LLM axes are sensitive to \emph{interpretable} linguistic content. We compiled a collection of features that together span multiple levels of linguistic analysis: morphology, syntax, and lexical semantics (Fig.~\ref{fig:overview}, bottom row). If these linguistic features align preferentially with the native axes of LLMs, we expect the matching score with each linguistic feature to decrease as $\alphaval$ increases.

Three of the feature sets are categorical minimal pairs in which one member is well-formed and the other is not: the \textit{Wug} test for morphological generalization (\citealp{berko1958child}; data from \citealp{hanna2026formal}); \textit{BLiMP} for syntax (\citealp{warstadt2020blimp}; one sub-task from each of BLiMP's 12 broad phenomenon classes); and \textit{hypernymy} for lexical category structure (\citealp{hanna2021analyzing}; see Fig.~\ref{fig:overview}, bottom row for examples of the minimal pairs). The remaining two feature sets are graded, multi-dimensional word-level norms: the Lancaster sensorimotor norms (\citealp{lynott2020lancaster}; ratings for six perceptual dimensions per word, henceforth \textit{perceptual}) and the \textit{affective norms} (\citealp{warriner2013norms}; valence, arousal, dominance), with a shared set of 13,372 words present in both norm sets. For the categorical tasks, the target $Y$ is a single binary variable indicating if the input is well-formed, and the matching score is a point-biserial correlation between each LLM unit and that label. For the rating tasks, the matching score is computed per rating dimension (the best-correlated LLM unit is allowed to differ across dimensions) and averaged over the six (perceptual) or three (affective) dimensions. Further details on all five feature datasets are given in the Methods.

For every feature category, alignment decreases as $\alphaval$ increases (Fig.~\ref{fig:2}; Stouffer \textit{Z} between 10.0 and 14.4, all \textit{p} < 10\textsuperscript{-22}). The mean difference between native and random is positive for every task (Wug: 0.038; BLiMP: 0.018; Hypernymy: 0.046; Perceptual: 0.054; Affective: 0.032), and across models, the native basis is more aligned than a random rotation for 75--100\% of models depending on the task (12/12 for BLiMP; 11/12 for Wug and Perceptual; 10/12 for Hypernymy; 9/12 for Affective; all Holm-corrected \textit{p} $\leq$ 0.016; see SI~\ref{sec:si-blimp} for results in all BLiMP sub-tasks). The privileged basis is therefore not tied to one specific level of linguistic analysis: it emerges for morphology, syntax, lexical-semantic categories, and graded dimensions of meaning.

\subsection{The native basis is far from optimal}

\begin{figure}[t]
  \centering
  \includegraphics[width=\textwidth]{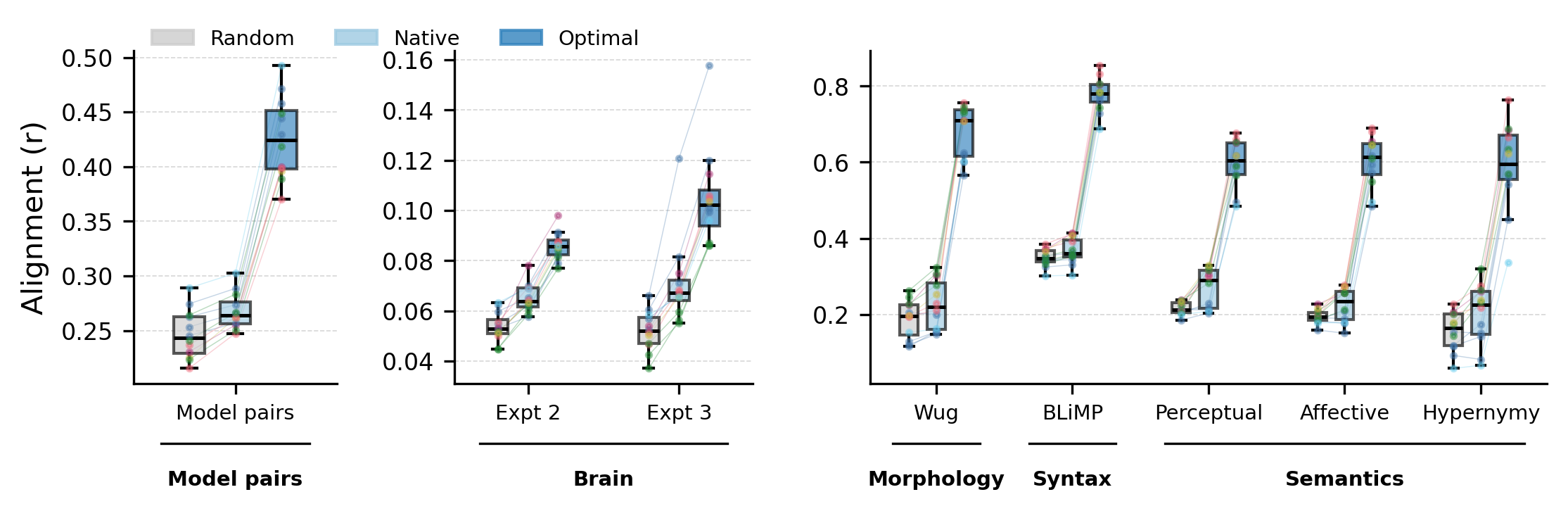}
  \caption{\textbf{Random, native and optimal alignment.} From left to right: \emph{model-to-model alignment} (with Pythia-160M as the reference); \emph{model-to-brain alignment} (Pereira Experiments 2 and 3); and \emph{model-to-feature alignment} (grouped by morphology, syntax, and lexical semantics). Each box shows the distribution across models of the alignment under a full random rotation ($\alpha = 1$, gray), the native basis ($\alpha = 0$, light blue), and the optimal axis alignment ($\Qstar$, dark blue); thin lines connect the three values of one model. Across all alignment targets, the pattern is \emph{random} $<$ \emph{native} $\ll$ \emph{optimal}.}
  \label{fig:3}
\end{figure}

The previous sections compared the native basis only against random uniformly drawn rotations of it. To set an upper bound on how much alignment can be recovered by any transformation that re-arranges the axes, we additionally compute the \emph{optimal axis alignment} score: for each comparison we learn on the training data an orthogonal $\Qstar$ that maximizes the matching score, and evaluate it on the same held-out data as the native basis (Methods). This optimal rotation gives us a ceiling for the pairwise matching score. 

Across all three settings, LLM representations in the native basis are more aligned with the target features than their random rotations, but substantially less aligned than what is achieved under the optimal rotation $\Qstar$ (\emph{random} $<$ \emph{native} $\ll$ \emph{optimal}, see Fig.~\ref{fig:3}; optimal exceeds native for 12/12 models on every target, all Wilcoxon \textit{p} = 5 $\times$ 10\textsuperscript{-4}). The substantial gap between native and optimal is not due to finding a specific optimal rotation for each feature: fitting a single shared orthogonal rotation jointly across all 16 feature tasks reproduces the scores in Fig.~\ref{fig:3} to within noise (\ref{sec:si-joint}). 

\subsection{The native basis supports sparse readouts}

\begin{figure}[t]
  \centering
  \includegraphics[width=\textwidth]{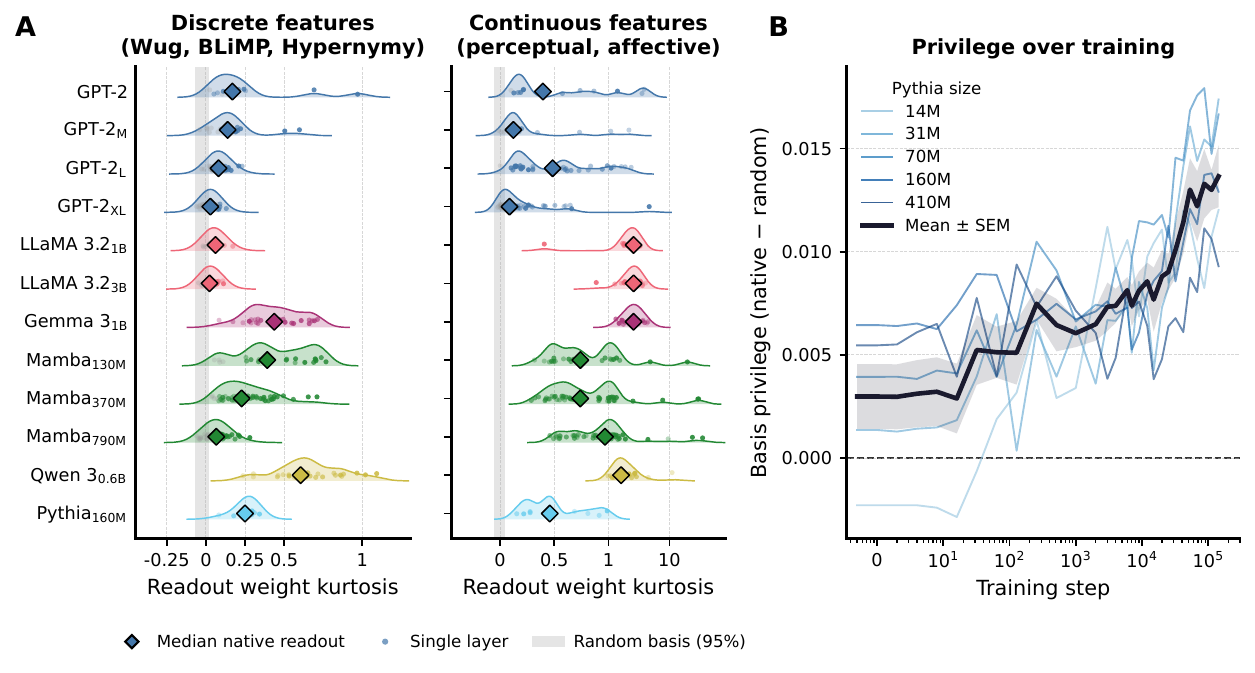}
  \caption{Readout sparsity and training dynamics. \textbf{A.} Kurtosis of the ridge readout weights across hidden units in the native basis, for the discrete features (left; Wug, BLiMP and hypernymy labels) and the continuous features (right; perceptual and affective norms). 
  For each model, the line shows the distribution over layers of the native kurtosis (median over tasks); the small dots are the individual layers (darker = deeper; gray if within the null of that layer), and the vertical square is the median over tasks and layers. The gray band is the 95\% interval of this median under a full random rotation of the basis ($\alpha = 1$, 10 rotation samples) using the widest interval across models. \textbf{B.} Basis privilege across training. The difference between native and random alignment across 29 log-spaced Pythia training checkpoints, for five model sizes. The thick line indicates the mean across sizes; the shaded band around it, the SEM across sizes. The basis privilege is positive but small at random initialization and increases with training.}
  \label{fig:4}
\end{figure}

Now that we have shown that a partially canonical representation basis exists in trained LLMs, we next ask what distinguishes this basis from its rotations. What are the consequences of this non-arbitrary representation basis, and what are its computational advantages? One possibility is that the native coordinate system supports sparse coding. Sparse codes, in which each downstream computation reads from only a few units, have received great attention in neuroscience, and have been argued to minimize interference between simultaneously represented features \citep{kanwisher2010functional, khosla2024privileged}. A sparse code is also \emph{basis-dependent}: the units along which the code is sparse define a privileged coordinate frame, and rotating away from that frame re-organizes the same information through a denser representation, possibly more prone to interference.

We test this idea by measuring \emph{readout weight sparsity}: whether the downstream readout of a feature is concentrated on a small subset of hidden units. For each combination of task, model, and layer, we fit a ridge regression from the hidden activations (z-scored per unit) to the task target (the binary acceptability label for Wug, BLiMP and hypernymy; the vector of perceptual and affective norms for the word stimuli) and compute the Fisher kurtosis of the regression weights across hidden units: high kurtosis means that a few units carry almost all of the readout weight, low kurtosis that the weight is spread approximately evenly across units.

Readouts are sparser in the native basis than random rotations in every model on both the discrete and the continuous features (Fig.~\ref{fig:4}A). Thus, in its native basis, a model reads each of these features from a small subset of units to an extent that no random rotation of the same representation can reproduce.

\subsection{The privileged basis strengthens through training}

The remaining question is \emph{when} the privileged basis emerges: is it built-in into the architecture at random initialization, or is it learned during training?

To answer this question, we monitor the basis privilege (the difference in pairwise alignment between the native and the randomly rotated basis) across 29 log-spaced training checkpoints of the Pythia suite \citep{biderman2023pythia, van2025polypythias}, at five model sizes (14M, 31M, 70M, 160M, 410M parameters). To analyze the emergence dynamics of the basis privilege we use again the model-to-model approach: at each checkpoint, we compute the matching score between two models of the same size, but trained with a different random seed and data order.

Two observations emerge from Fig.~\ref{fig:4}B. First, at random initialization (step 0, where both models have only their architectural priors and no exposure to language) the basis privilege is positive for four of the five sizes, but small and not significant (range --0.002 to 0.006; mean 0.003, 95\% CI [--0.001, 0.007]; \textit{t}(4) = 1.9, \textit{p} = 0.13). This shows that an architectural contribution to the privileged basis, if present, is small.

Second, the gap increases through training in every model size (Spearman $\rho$ between log step and gap between 0.42 and 0.95, all \textit{p} < 0.03), reaching about 0.009 to 0.017 by 143,000 steps (mean 0.014, 95\% CI [0.010, 0.018]; the difference between the final and the initial gap is positive for 5/5 sizes, \textit{t}(4) = 4.5, \textit{p} = 0.01). The largest increase happens late in training (after roughly 30,000 steps, i.e., about 60 billion tokens), and continues until the last checkpoint. This trajectory highlights a substantial \emph{learned} component that accumulates throughout pre-training. 

\section{Discussion}

An important question to understand how LLMs represent language is whether the native coordinate axes of their representations are meaningful or whether they are interchangeable with any other basis. Across three alignment targets, we consistently found that the native basis is more aligned than a random rotation (in line with prior work showing that transformer residual streams may have a privileged basis; \citealp{elhage2023privilegedbasis, adamprivilegedbasis}), yet clearly less aligned than the best achievable orthogonal transformation (a large gap for model and feature targets, a smaller one for the brain data). LLM coordinates are therefore neither fully arbitrary nor fully canonical, but \emph{partially} canonical. This finding implies that the contrast between unit-based and population-based views outlined in the introduction is an oversimplification. The native axes of an LLM are thus meaningful in their own right, but examining units in isolation gives only a partial account of how the model organizes its representations. This pattern emerged in standard transformer models and also in Mamba, a state-space model, and the overall pattern was very similar across them. The partly canonical basis is therefore not specific to self-attention but a broader feature of modern sequence models trained on language. Our findings therefore challenge theories of representational convergence to explain both why language models learn similar representational geometries and why some of that shared structure emerges preferentially in the tuning of individual units.

A core feature of the native basis appears to be the possibility to decode linguistically meaningful information through \emph{sparse readouts}: downstream computations can recover useful variables from a small subset of native units, and rotating away from this basis forces the same information through a denser code. A sparse readout may be computationally useful because it limits interference at two levels. During processing, representing many kinds of information at once is easier if their codes remain separable; codes that are heavily intermixed are more likely to collide and degrade one another \citep{reddy2007category, kanwisher2010functional, musslick2017multitasking}. During learning, when the units supporting different computations overlap, an update that helps one type of computation tends to perturb the others---the kind of interference long recognized as a fundamental problem for learning in distributed systems \citep{mccloskey1989catastrophic, french1999catastrophic}. Concentrating each computation on a few units could protect against both: information can be represented separately, and one function can be updated without overwriting another. Yet the native basis in LLMs is substantially less aligned with our target features than the best orthogonal rotation, which suggests that reducing interference is not the only pressure shaping model representations. Language models must also encode many features within hidden states of fixed size, and a fully canonical code with one axis for each variable would plausibly sacrifice too much flexibility. Partially canonical axes may therefore reflect a compromise: information is concentrated enough to reduce interference, yet superposed enough to fit within a limited dimensionality, balancing interference minimization against expressivity. Note, however, that sparsity does not by itself trivially imply reduced interference; this is a consequence of the sparse code that we did not test directly and that remains open for future work.

What do our findings tell us about which interpretability tools are licensed? Recent work in mechanistic interpretability has increasingly focused on identifying low-dimensional structure within high-dimensional LLM activation spaces, such as subspaces and curved manifolds (e.g., days of the week on a circle, numbers on a helix; \citealp{engels2025not,kantamneni2025language}). These geometric structures are now proposed as units of analysis, both for discovering interpretable features \citep{bhalla2026sae,fel2026structuring} and for steering behavior \citep{wurgaft2026manifold}. 
In line with the \textit{directional} view presented in the introduction, these methods typically rely on distances and inner products of the activations, making them unaffected by rotations of the basis. Our findings show that there is key information to be gained from native space analyses (which rotation-invariant analyses miss), but also that the native basis is far from optimal. 
Of course, this does not undermine the structures recovered by geometric methods. 
The large gap between native and optimal alignment suggests that most target-relevant structure is not axis-aligned, supporting the use of population-level, geometric methods that might capture such structure. That said, even in the geometric analyses, individual units can serve as local readouts of a manifold (e.g., curve-detector neurons; \citealp{cammarata2020curve,fel2026structuring}).

A second recent trend in mechanistic interpretability returns to the native units of networks, with a strong focus on MLP neurons within each transformer block \citep{gurnee2024universal,shabalin2026comparing}, rather than the residual stream that we study here. For example, circuits traced through MLP neurons are as sparse as those built from sparse autoencoder features \citep{arora2026language}, and it has been argued that neurons and circuits should be studied directly as they are already ``clean'' and interpretable \citep{cammarata2026clean}.
Our results provide complementary evidence for studying the native basis, but differ from it in two key ways. First, in MLPs the element-wise nonlinearity makes the neuron basis privileged by construction \citep{elhage2022toy} and no such nonlinearity is performed on the residual stream neurons (see below). Second, a privileged axis does not need to be monosemantic. We ask whether a unit carries more target-relevant information than a random direction, not whether it encodes a single (interpretable) feature. 


Why do similar patterns of tuning recur along individual coordinates of the residual stream in independently trained language models? The \textit{contravariance principle} \citep{cao2021explanatory} states that harder tasks have fewer possible solutions, so networks trained on the same hard tasks tend to converge on similar solutions.  \citet{yamins2026contravariance} formalized this idea and showed that this convergence extends to single units when non-linearities are applied element-wise, as in transformer MLPs. Residual coordinates do not have element-wise non-linearities, and \citeauthor{yamins2026contravariance}'s account therefore does not predict a privileged basis in the residual stream. However, our results show that such a preference exists. At random initialization the basis privilege is close to zero but slightly positive (Fig.~\ref{fig:4}B), so an architectural contribution cannot be excluded; however, most of the privilege emerges through training (Fig.~\ref{fig:4}B). One possible source is Adam, which scales the update of each weight separately \citep{adamprivilegedbasis}. This includes the MLP output weights, which determine how much each MLP unit contributes to each dimension in the residual stream \citep{elhage2023privilegedbasis}. If a dimension in the residual stream receives most of its input from a small number of MLP units, it may inherit their tuning, and given that MLP units can converge across models, so would the residual stream. Future work could test whether privileged dimensions in the residual stream are driven by MLP units that are also privileged.

Privileged axes are also found in human perceptual cortices and convolutional neural networks, with one substantial difference: the native basis is much closer to the optimal alignment than in our case \citep{khosla2024privileged}. Whether this difference is specific to current language models or reflects something about language itself is an open question. One possibility is that language requires strong representational reuse: the same hidden state must support multiple dimensions of linguistic structure and meaning---from morphology and syntax to lexical semantics, discourse, and reasoning. That pressure may affect the tradeoff between minimizing interference and preserving expressivity, moving the code away from a fully canonical basis. Evidence from neuroscience suggests that linguistic computations in the human brain are also characterized by mixed selectivity, where lexical meanings and syntactic processes are supported by the same brain regions (the whole language network: \citealp{fedorenko2020lack, shain2024distributed}, cf. \citealp{dighiero2026distinct}). The strongly privileged axes of sensory cortices and vision networks may thus be one end of a gradient along which language sits closer to mixed selectivity, which would account for the more subtle effects we find here.

Taken together, these findings suggest that LLM representations may fall between two extremes: they are not an arbitrary coordinate system nor a collection of interpretable axes, but a partially canonical code that affords a sparse readout of meaningful information.

\section{Methods}

\subsection{Stimuli and alignment targets}

\subsubsection{Pereira fMRI dataset} The brain data are taken from Experiments 2 and 3 of \citet{pereira2018toward}, which together comprise fMRI recordings from 10 unique participants (9 in Experiment 2, 6 in Experiment 3; 5 participants in both) reading 627 sentences (384 in Experiment 2, 243 in Experiment 3), grouped into passages of three to four sentences each. The two sets of materials spanned a broad range of content areas; sentences were 7--18 words long in Experiment 2 and 5--20 words long in Experiment 3. Sentences were presented visually one at a time for 4 seconds, followed by 4 seconds of fixation, with an additional 4 seconds of fixation at the end of each passage. Each participant saw each sentence between 1 and 4 times, with the majority seeing each sentence 3 times. We average each voxel's responses over the repetitions of each sentence, and exclude the one participant (in Experiment 2) who saw each sentence only once, leaving 9 unique participants (8 in Experiment 2, 6 in Experiment 3; 5 in both). We use the participant-level responses in the language network, defined functionally in each participant using the standard language localizer \citep{fedorenko2010new}. We restrict our analyses to the top-10\% language-selective voxels in each participant (1,364 voxels) and, within these, to the 20\% (273 voxels) with the highest test-retest reliability: the Pearson correlation across sentences between a voxel's responses in two sessions, averaged over all pairs of sessions. Reliability is calculated from the brain data alone so the selection does not consider representations from language models; see SI~\ref{sec:si-voxsel} for results with all language voxels. Because the sentences of a passage were presented consecutively and share both content and slow BOLD autocorrelation, all brain alignment scores are estimated with passage-level 10-fold cross-validation. Unit matching (and the optimal rotation) is fitted on the training passages and scored on the held-out passages, per voxel; scores are averaged over folds, voxels, and the 10 rotation samples. Additional details on the fMRI analyses can be found in SI~\ref{sec:si-fmri}.

\subsubsection{Linguistic feature datasets} We consider five datasets that span different levels of linguistic analysis: morphology, syntax and lexical semantics. Three are binary minimal pair tasks (one well-formed item, one ill-formed item, predict a binary $\{0,1\}$ acceptability label); two are continuous lexical norm tasks (predict a multi-dimensional rating vector for each word).

\textbf{\emph{Wug} (morphology).} The \textit{Wug} test (\citealp{berko1958child}; data from \citealp{hanna2026formal}) probes morphological generalization with 998 pseudoword minimal pairs differing in a single inflectional ending (we removed two pairs where the alternatives were identical). The model must distinguish well-formed from ill-formed inflection of a novel stem.

\textbf{\emph{BLiMP} (syntax).} The Benchmark of Linguistic Minimal Pairs for English \citep{warstadt2020blimp} contains 67 minimal-pair sub-tasks grouped into 12 broad linguistic-phenomenon classes; we sample one sub-task per class, giving the 12 sub-tasks listed in Table~\ref{tab:blimp} (1,000 minimal pairs per sub-task, 12,000 pairs total).

\begin{table}[t]
  \centering
  \small
  \begin{tabular}{ll}
    \toprule
    \textbf{Phenomenon class} & \textbf{Sub-task sampled} \\
    \midrule
    Anaphor agreement          & \textsc{anaphor\_gender\_agreement} \\
    Argument structure         & \textsc{transitive} \\
    Binding                    & \textsc{principle\_A\_c\_command} \\
    Control / raising          & \textsc{existential\_there\_object\_raising} \\
    Determiner--noun agreement & \textsc{determiner\_noun\_agreement\_1} \\
    Ellipsis                   & \textsc{ellipsis\_n\_bar\_1} \\
    Filler--gap dependency     & \textsc{wh\_questions\_subject\_gap} \\
    Irregular forms            & \textsc{irregular\_past\_participle\_adjectives} \\
    Island effects             & \textsc{adjunct\_island} \\
    NPI licensing              & \textsc{matrix\_question\_npi\_licensor\_present} \\
    Quantifiers                & \textsc{existential\_there\_quantifiers\_1} \\
    Subject--verb agreement    & \textsc{irregular\_plural\_subject\_verb\_agreement\_1} \\
    \bottomrule
  \end{tabular}
  \caption{The 12 BLiMP sub-tasks used in the model-to-feature
  alignment analyses, one per broad linguistic-phenomenon class.}
  \label{tab:blimp}
\end{table}

\textbf{\emph{Hypernymy} (lexical semantics).} 551 minimal pairs adapted from \citet{hanna2021analyzing}, contrasting a true hypernym with a category-incompatible distractor in a frame of the form ``Consider X, and other \textsc{noun}'' (e.g., \textit{``Consider caterpillars, and other insects/gems''}).

\textbf{\emph{Perceptual norms} (lexical semantics).} The Lancaster sensorimotor norms \citep{lynott2020lancaster} provide continuous ratings of perceptual strength on six sensory dimensions (visual, auditory, haptic, gustatory, olfactory, interoceptive) for 39,707 English words.

\textbf{\emph{Affective norms} (lexical semantics).} \citet{warriner2013norms} provide continuous valence, arousal and dominance ratings for 13,915 English lemmas. These three dimensions describe the degree to which a word's referent is experienced as pleasant rather than unpleasant (valence), as activating rather than calming (arousal), and as eliciting a feeling of control over rather than being controlled by the referent (dominance). For instance, \say{calm} is pleasant but low-arousal ($V = 6.9$, $A = 1.7$, $D = 7.4$), \say{panic} is unpleasant, highly arousing and low in control ($V = 2.6$, $A = 6.4$, $D = 2.7$), and \say{triumph} is pleasant, arousing and high in control ($V = 7.3$, $A = 5.6$, $D = 7.5$). For the perceptual and affective tasks we restrict to a shared set of 13,372 words present in both norm sets, and present each word in the template ``The meaning of the word `\textit{word}'\,'' to the LLMs. For each rating dimension, the matching score finds the best-correlated LLM unit independently; the per-task score is the mean over the six (perceptual) or three (affective) dimensions.

\subsection{Activation extraction}

For each model and stimulus set we extract activations in the residual stream at every transformer block. For each sentence we average the token activations, excluding the first token and, when it is punctuation or a special token, the last token (we found that these tokens had an activation norm orders of magnitude larger than the other tokens across most models, which inflated our results; see also \citealp{xiao2024efficient, sun2024massive}). For the word-level semantic stimuli we average only across the tokens of the target word. All analyses in the main text are reported for the residual stream after each transformer block, averaging across layers (but excluding the embedding layer). 

\subsection{Pairwise matching score}

We use the pairwise matching score introduced by \citet{khosla2024privileged}. Consider two matrices $X \in \mathbb{R}^{n \times d_X}$ and $Y \in \mathbb{R}^{n \times d_Y}$ with the activations of $d_X$ source units (the rotated LLM representation) and $d_Y$ target units (the alignment target) to the same $n$ stimuli; $X_i$ and $Y_j$ indicate the responses of source unit $i$ and target unit $j$ across the $n$ stimuli. We proceed in two stages. First, on a training split, we match each target unit $j$ to the single source unit $i$ whose response profile is most correlated with it in absolute value, so that axes that are negatively correlated count as matches. We encode the matching with binary indicators $a_{ij} \in \{0, 1\}$, where $a_{ij} = 1$ means source unit $i$ is assigned to target unit $j$ and $a_{ij} = 0$ means it is not; the constraint $\sum_i a_{ij} = 1$ for every $j$ forces each target unit to be matched to exactly one source unit. The optimal matching maximizes the total correlation of the assigned pairs,

\begin{equation}
  \{a^*_{ij}\}_{i = 1, \ldots, d_X;\; j = 1, \ldots, d_Y}
  = \operatorname*{argmax}_{a_{ij}}
    \sum_i \sum_j \bigl|\rho\!\left(X^{\mathrm{train}}_i,\, Y^{\mathrm{train}}_j\right)\bigr| a_{ij}
  \label{eq:matching}
\end{equation}
where $\rho(\cdot,\cdot)$ is the Pearson correlation over the training stimuli. Second, we fix this matching $a^*$ and evaluate it on a held-out test split: for each target unit we take its test-set Pearson correlation with the source unit it was matched to, and average these correlations over all $d_Y$ target units, 
\begin{equation}
  s_{\text{pairwise}}(X^{\mathrm{test}}, Y^{\mathrm{test}}, a^*)
  = \frac{1}{d_Y} \sum_i \sum_j \rho\!\left(\sigma_{ij}X^{\mathrm{test}}_i,\, Y^{\mathrm{test}}_j\right) a^*_{ij}.
  \label{eq:pairwise}
\end{equation}
Where $\sigma_{ij} = \operatorname{sign}\,\rho(X^{\mathrm{train}}_i, Y^{\mathrm{train}}_j)$ is the sign of the correlation. The score is asymmetric by design. The semi-matching constraint (each $Y$ unit is assigned one $X$ unit; multiple $Y$ units may be assigned the same $X$ unit) accommodates representations of different width. This type of alignment is known as bipartite semi-matching in graph theory \citep{li15convergent, lawler2001combinatorial}. Train/validation/test splits use a 60/20/20 partition of the stimuli for the model-to-model, model-to-feature and training-dynamics analyses (for the minimal pair tasks we design the splits so that the two members of a pair always appear in the same split to avoid leakage), and passage-level 10-fold cross-validation for the brain data; further details are in SI~\ref{sec:si-splits}.

\subsection{Axis rotation}

To probe for axis dependence, we rotate $X$ by orthogonal matrices. Let $Q \sim \mathrm{Uniform}(\SOnn)$. Following \citet{khosla2024privileged}, we construct fractional rotations $Q^{\alpha}$ for $\alpha \in [0, 1]$ via Schur decomposition: writing $Q = W T W^*$ with $T$ upper quasi-triangular, the fractional power is $Q^{\alpha} = W T^{\alpha} W^*$, where $T^{\alpha}$ is computed block-wise on the $2 \times 2$ blocks of $T$. We sample 10 such rotation matrices per condition and average over them; for each sample we evaluate at 11 values of $\alpha$ ($\alpha = 0$ plus 10 evenly-spaced values in $[0.1, 1]$). $\alpha = 0$ is the identity (native basis); $\alpha = 1$ is a fully random rotation.

\subsection{Optimal axis alignment}

To find the maximum alignment achievable under any orthogonal rotation of $X$, we solve
\[
  \Qstar = \operatorname*{argmax}_{Q \in \SOnn} s_{\text{pairwise}}(XQ, Y).
\]
Directly optimizing over $\SOnn{}$ is problematic because we would have to keep $Q$ orthogonal at every step. We avoid the constraint by reparametrising: every rotation matrix can be written as $Q = \exp(S)$ for some skew-symmetric matrix $S$ (i.e.\ $S^\top = -S$), where $\exp(\cdot)$ is the matrix exponential. Optimizing over $S$ instead of $Q$ is unconstrained -- any $S$ with $S^\top = -S$ produces a valid rotation $Q$ -- so we can use a standard gradient-based optimizer and backpropagate through the matrix exponential. This is the exponential map trivialization of \citet{lezcano2019trivializations} as implemented in GeoTorch; we use its static form with the base point fixed at the identity, so that optimization starts from $S = 0$, i.e.\ $Q = I$, the native basis. We use Adam with learning rate $10^{-4}$ and a one-cycle cosine schedule peaking at $10^{-3}$, for 50 epochs. The best epoch is selected by the matching score on a held-out validation split; the test-split matching score at that epoch is the optimal axis alignment. The objective uses the same matching as the pairwise score, so it is invariant to sign. Remaining hyperparameters are in SI~\ref{sec:si-splits}.

\subsection{Statistical analysis}

The unit of analysis is the model. For each alignment target we obtain for each model one value per condition (native, random with $\alpha = 1$, optimal) by averaging over layers above the embedding layer (see SI~\ref{sec:si-layers} for robustness across different layers) and, where applicable, over participants (brain) or sub-tasks (BLiMP), so that \textit{n} = 12 models. The basis privilege (native -- random) and the ceiling gap (optimal -- native) are tested against zero with exact two-sided Wilcoxon signed-rank tests, with Holm's correction across the eight alignment targets for the comparison of native vs. random. The decline of alignment with $\alpha$ is tested for each model with Kendall's $\tau$ between the $\alpha$ value and the alignment value (exact one-sided \textit{p}), and the \textit{p}-values are combined across models with Stouffer's \textit{Z}. The code for all analyses is available on \href{https://github.com/Andrea-de-Varda/partially-canonical-axes}{GitHub}.

\subsection{Readout sparsity}

For each model, layer and task we fit a ridge readout ($\lambda = 1$; a ridge classifier for the minimal pair tasks and a ridge regression on the nine perceptual and affective dimensions) from the activations to the task target and take the Fisher kurtosis of its weights across units as a measure of readout sparsity. Before fitting the readout we z-scored each unit so that the kurtosis reflects how the readout weight is distributed across units and not differences in their scale. We constructed an empirical null distribution of the kurtosis of a model's ridge weights from 10 random rotations with $\alpha = 1$: for each rotation, we take the median kurtosis from the same layers and tasks as in the native basis; the median kurtosis is above the null when its value in the native basis is higher than the mean plus 1.96 standard deviations of these 10 values. 

\subsection{Training dynamics analysis}

For the analysis of the training dynamics we use the Pythia suite \citep{biderman2023pythia, van2025polypythias}, which provides 154 training checkpoints for each model size. We select 29 log-spaced steps (geometric spacing from step 1 to step 143{,}000, approximated to the nearest available checkpoint, plus step 0 and step 143{,}000). For each selected step and each of the five model sizes, we extract residual-stream activations from two models only differing in their training seed in response to the combined Pereira stimuli. Then, we compute the pairwise matching score at the 11 values of $\alpha$. In the main text (Fig.~\ref{fig:4}B) we report only the difference between native and random, which indicates the basis privilege; the underlying absolute trajectories of native, random and optimal axis alignment per size, together with a rotation-invariant RSA control, are reported in SI~\ref{sec:si-trajectories} and Fig.~\ref{fig:S1}.

\section*{Acknowledgements}

A.G.d.V. was supported by the K. Lisa Yang ICoN Center Postdoctoral Fellowship and also acknowledges a Gemini Academic Program Award. G.T. acknowledges support from the Chan Zuckerberg Initiative Foundation to establish the Kempner Institute for the Study of Natural and Artificial Intelligence at Harvard University. E.F. was supported by National Institutes of Health award U01-NS121471 and funds from the McGovern Institute for Brain Research, the Brain and Cognitive Sciences Department, the Simons Center for the Social Brain and the MIT Siegel Family Quest for Intelligence. M.K. was supported by a Coefficient Giving (formerly Open Philanthropy) Grant. 

\section*{Author contributions}
\textbf{A.G.d.V.}: Conceptualization, Methodology, Software, Formal analysis, Investigation, Data curation, Validation, Visualization, Project administration, Funding acquisition, Writing (original draft), Writing (review and editing).
\textbf{F.D.}: Software, Formal analysis, Validation.
\textbf{G.T.}: Writing (original draft), Writing (review and editing).
\textbf{E.F.}: Conceptualization, Supervision, Funding acquisition, Writing (review and editing).
\textbf{M.K.}: Conceptualization, Supervision, Funding acquisition, Writing (review and editing).

\subsection*{Use of AI tools}
Claude Opus 4.8--5.5 (Anthropic) was used for coding assistance during the analyses and for writing assistance. Its role in the Introduction, Results and Discussion was minor and limited to editing the authors' text and improving cohesiveness. It played a larger role in the writing of the Methods and the Supplementary Information. All ideas, conceptual choices and decisions about study design, analyses and interpretation are the authors' own. All code and text were reviewed and verified by the authors, who take full responsibility for the content of this article.

\bibliography{refs}

\clearpage
\setcounter{figure}{0}
\renewcommand{\thefigure}{S\arabic{figure}}
\setcounter{section}{0}
\renewcommand{\thesection}{S\arabic{section}}

\titleformat{\section}[hang]
  {\Large\bfseries\color{accent}}
  {\thesection.}{0.5em}
  {}
  [\vspace{-3pt}{\color{rulegrey}\titlerule[0.6pt]}]

\section*{\color{accent}\Large Supplementary Information}

\section{Joint optimal rotation across feature tasks}
\label{sec:si-joint}

The gap between the alignment obtained in the native and in the optimal basis (Fig.~\ref{fig:3}) might be inflated by the way optimal axis alignment is computed: each feature task fits its own $Q^*$, whereas the native basis is shared across all tasks. The concern is that there might be no single rotation that simultaneously outperforms the native basis on all tasks, and the optimal rotation for each task would select different best axes for different targets.

We test this by fitting a single shared rotation jointly across all 16 feature tasks. For each combination of model and layer, we minimize the sum (unweighted) of task-level pairwise-matching losses on the training splits, with the same 60/20/20 train/validation/test split and the same validation-based early stopping. The  joint score reproduces the task scores to within noise: averaged across models, the joint optimal axis alignment is 0.679 for Wug (per-task 0.678), 0.782 for BLiMP (per-task 0.778), 0.583 for perceptual ratings (per-task 0.598), 0.598 for affective ratings (per-task 0.603), and 0.589 for hypernymy (per-task 0.590). The largest absolute difference is 0.015. A single coordinate system in the source model can therefore be rotated to align with all feature targets at once.

\section{Absolute training trajectories and RSA control}
\label{sec:si-trajectories}

\begin{figure}[t]
  \centering
  \includegraphics[width=\textwidth]{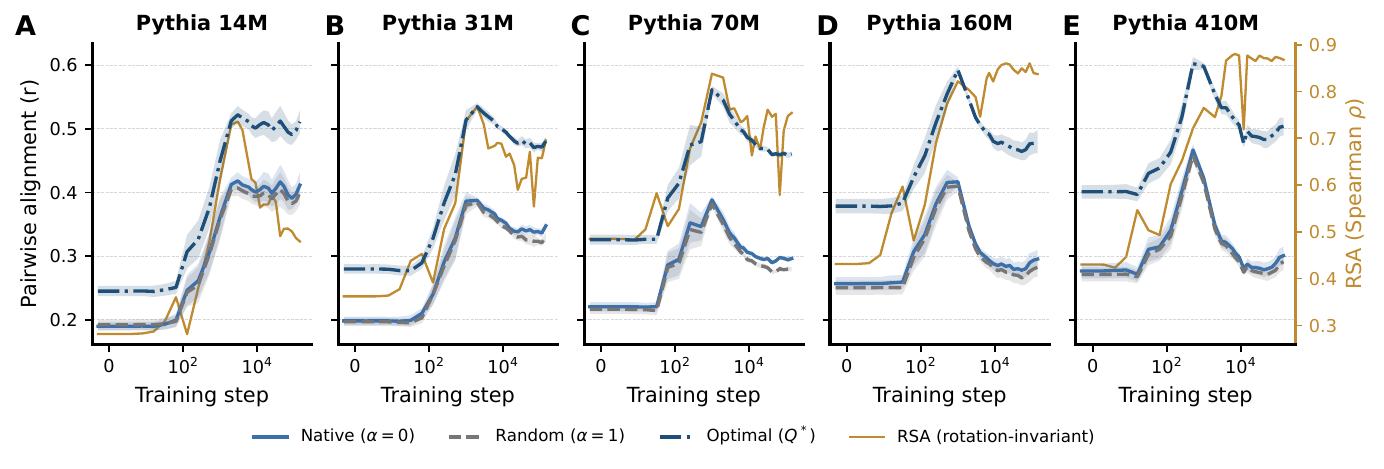}
  \caption{\textbf{Training trajectories across sizes and rotation-invariant RSA control.} Each panel shows the trajectories of native (blue, solid), random (gray, dashed) and optimal axis alignment (dark blue, dash-dotted) for one Pythia size, as a function of training step (log scale); shaded areas are $\pm 1$ SEM across layers. The privilege gap (Fig.~\ref{fig:4}B) is the difference between native and random. The ochre line (right axis) is the RSA control across the same checkpoints; RSA peaks and declines in the small models but increases in the larger ones.}
  \label{fig:S1}
\end{figure}

Fig.~\ref{fig:4}B in the main text reports only the privilege \emph{gap} (native -- random). The underlying absolute trajectories of native, random and optimal axis alignment differ qualitatively across model sizes (Fig.~\ref{fig:S1}). In the smallest model (14M), all three quantities rise rapidly and then plateau. In the larger models (31M, 70M, 160M, 410M), all three rise to a peak around step 500--2{,}000 (roughly $10^9$ tokens), then drop and recover partially. The privilege gap (the difference between native and random) increases slowly even as the absolute alignment values fluctuate dramatically.

To check that this pattern is about coordinate-axis alignment (vs. overall representational similarity), we compute an axis-insensitive control alongside the matching score. Representational similarity analysis (RSA; \citealp{kriegeskorte2008representational}) compares the geometry of pairwise stimulus similarities and is invariant to orthogonal rotations of the underlying basis, in contrast to the axis-dependent metric we use elsewhere. The two metric families show different patterns (Fig.~\ref{fig:S1}). In small models RSA peaks and then declines, but in large models (160M, 410M) RSA increases throughout training even as the pairwise matching score peaks and drops.

\section{fMRI analyses}
\label{sec:si-fmri}

This section documents the fMRI acquisition, preprocessing and functional-region-of-interest (fROI) definition procedure that produced the participant-level language-network responses we use as the brain alignment target in Section ``Model-to-brain alignment''.

\subsection{Acquisition}

Structural and functional MRI data were acquired on a whole-body 3 Tesla Siemens Trio scanner with a 32-channel head coil at the Athinoula A. Martinos Imaging Center at the McGovern Institute for Brain Research at MIT. T1-weighted structural images were collected in 179 sagittal slices with 1\,mm isotropic voxels (TR\,=\,2{,}530\,ms, TE\,=\,3.48\,ms). Functional, blood-oxygenation-level-dependent (BOLD) data were acquired with an EPI sequence (90$^{\circ}$ flip angle; GRAPPA with an acceleration factor of 2): thirty-one 4\,mm near-axial slices in interleaved order with a 10\% distance factor; 2.1\,mm $\times$ 2.1\,mm in-plane resolution; 200\,mm field of view in the phase-encoding (anterior-to-posterior) direction; matrix size 96\,$\times$\,96; TR\,=\,2{,}000\,ms; TE\,=\,30\,ms. Prospective acquisition correction adjusted gradient positions based on participant motion one TR back. The first 10\,s of each run were discarded to allow for steady-state magnetization.

\subsection{Preprocessing}

Preprocessing was performed in SPM12 with default parameters except where noted, using custom MATLAB scripts. Anatomical data were normalized into MNI space and segmented into probabilistic maps of gray matter (GM), white matter (WM) and cerebro-spinal fluid (CSF); a GM mask was generated from the GM probability map and resampled to 2\,mm isotropic voxels to mask the functional data. Functional data were motion-corrected (realignment to the mean image using second-degree b-spline interpolation) and normalized into MNI space (estimated on the mean functional image with trilinear interpolation) using SPM12's unified segmentation-and-normalization procedure with a reference image computed as the mean of the motion-corrected functional data after omitting outlier scans. The output was resampled to a common bounding box between MNI coordinates ($-90$, $-126$, $-72$) and ($90$, $90$, $108$), using 2\,mm isotropic voxels and 4th-degree B-spline interpolation, and smoothed spatially with a 4\,mm full-width-half-maximum Gaussian kernel.

\subsection{General Linear Model and language-fROI definition}

Effects were estimated using a General Linear Model (GLM) in which each experimental condition was modeled with a boxcar function convolved with the canonical hemodynamic response function (fixation was modeled implicitly, such that all time points outside an explicitly modeled condition were treated as fixation). Temporal autocorrelations in the BOLD time series were accounted for by a combination of high-pass filtering with a 128\,s cutoff and AR(0.2) whitening, approximating the observed covariance of the functional data under Restricted Maximum Likelihood estimation. The GLM design also included first-order temporal derivatives for each condition (to model variability in HRF delays) and nuisance regressors controlling for slow linear drifts, six subject-motion parameters, and outlier scans.

Language fROIs were defined in individual participants \citep{saxe2006divide, fedorenko2010new} by combining two sources of information: (i) the participant's activation map for a language localizer contrast, and (ii) group-level constraints (``parcels'') that delineated the expected gross location of each fROI while accommodating individual variability. The localizer is the English sentences-vs-nonwords contrast of \citet{fedorenko2010new}, in which blocks of intact English sentences are contrasted with blocks of length-matched nonword lists, with stimuli presented one (non)word at a time in a standard blocked design. The parcels were derived from a probabilistic activation overlap map of the sentences-vs-nonwords contrast in 220 independent participants, using watershed parcellation \citep{fedorenko2010new}, and cover extensive portions of the lateral frontal, temporal and parietal cortex. For each participant, five language fROIs were defined as the top-10\% of voxels with the highest $t$-values for the contrast within each parcel in the dominant (left) hemisphere: three on the lateral surface of the frontal cortex, and two on the lateral surface of the temporal and parietal cortex. All parcels are publicly available at \href{https://www.evlab.mit.edu/resources-all/download-parcels}{evlab.mit.edu/resources-all/download-parcels}.

\subsection{Repetition averaging, voxel selection and results with all voxels}
\label{sec:si-voxsel}

Each sentence was presented once per scanning session. Of the participants in our final sample, seven in Experiment 2 participated in three sessions and one in two; in Experiment 3, five participated in three sessions and one in four. For each voxel we average its responses to a sentence over sessions. To select voxels, we estimate each voxel's test-retest reliability as the Pearson correlation, across all sentences of an experiment, between its responses in two sessions, averaged over all pairs of sessions, and keep the 20\% of the 1,364 language-network voxels with the highest reliability (273 voxels per participant). The mean reliability of the selected voxels is 0.089 in Experiment 2 and 0.131 in Experiment 3 (0.030 and 0.056 across all voxels). The selection uses only the brain data and thus it cannot favor the native basis over its rotations. Nonetheless, for completeness, we ran the analyses on all language-responsive voxels, and the results are qualitatively the same. Alignment declines with $\alpha$ (Stouffer \textit{Z} = 13.3 and 14.5, both \textit{p} < 10\textsuperscript{-39}); the native basis is more aligned than a random rotation (native 0.044 vs. random 0.037 in Experiment 2; 0.047 vs. 0.036 in Experiment 3; a difference of 0.007 and 0.010, 95\% CI [0.005, 0.010] and [0.006, 0.015]; all models positive in each experiment, Wilcoxon \textit{p} = 5 $\times$ 10\textsuperscript{-4} in both); and the optimal axis alignment (0.057 and 0.067) is higher than the native alignment for all models in both experiments.

\section{Pairwise matching: data splits and hyperparameters}
\label{sec:si-splits}

\begin{table}[t]
  \centering
  \small
  \setlength{\tabcolsep}{1pt}
  \begin{tabular*}{\textwidth}{@{}l@{\extracolsep{\fill}} c r@{\extracolsep{0pt}\,}l @{\extracolsep{\fill}}c r@{\extracolsep{0pt}\,}l @{\extracolsep{\fill}}c r@{\extracolsep{0pt}\,}l @{\extracolsep{\fill}}c r@{\extracolsep{0pt}\,}l@{}}
    \toprule
    & \multicolumn{12}{c}{\textbf{Normalized depth}} \\
    \cmidrule(lr){2-13}
    \textbf{Target} & \multicolumn{3}{c}{0.25} & \multicolumn{3}{c}{0.5} & \multicolumn{3}{c}{0.75} & \multicolumn{3}{c}{1.0} \\
    \cmidrule(lr){2-4} \cmidrule(lr){5-7} \cmidrule(lr){8-10} \cmidrule(lr){11-13}
    & $\Delta r$ & \multicolumn{2}{c}{$n_{+}$} & $\Delta r$ & \multicolumn{2}{c}{$n_{+}$} & $\Delta r$ & \multicolumn{2}{c}{$n_{+}$} & $\Delta r$ & \multicolumn{2}{c}{$n_{+}$} \\
    \midrule
    Model-to-model  & \phantom{$-$}0.0216 & 12/12 & \makebox[1.3em][l]{$^{***}$} & \phantom{$-$}0.0216 & 12/12 & \makebox[1.3em][l]{$^{***}$} & \phantom{$-$}0.0215 & 12/12 & \makebox[1.3em][l]{$^{***}$} & \phantom{$-$}0.0196 & 12/12 & \makebox[1.3em][l]{$^{***}$} \\
    Brain, Exp.\ 2  & \phantom{$-$}0.0115 & 12/12 & \makebox[1.3em][l]{$^{***}$} & \phantom{$-$}0.0109 & 9/12 & \makebox[1.3em][l]{$^{**}$} & \phantom{$-$}0.0133 & 12/12 & \makebox[1.3em][l]{$^{***}$} & \phantom{$-$}0.0105 & 10/12 & \makebox[1.3em][l]{$^{**}$} \\
    Brain, Exp.\ 3  & \phantom{$-$}0.0193 & 12/12 & \makebox[1.3em][l]{$^{***}$} & \phantom{$-$}0.0178 & 12/12 & \makebox[1.3em][l]{$^{***}$} & \phantom{$-$}0.0228 & 12/12 & \makebox[1.3em][l]{$^{***}$} & \phantom{$-$}0.0085 & 10/12 & \makebox[1.3em][l]{$^{*}$} \\
    Wug             & \phantom{$-$}0.0186 & 8/12 & \makebox[1.3em][l]{} & \phantom{$-$}0.0384 & 11/12 & \makebox[1.3em][l]{$^{**}$} & \phantom{$-$}0.0282 & 10/12 & \makebox[1.3em][l]{} & \phantom{$-$}0.0839 & 12/12 & \makebox[1.3em][l]{$^{***}$} \\
    BLiMP           & \phantom{$-$}0.0130 & 9/12 & \makebox[1.3em][l]{$^{*}$} & \phantom{$-$}0.0161 & 11/12 & \makebox[1.3em][l]{$^{**}$} & \phantom{$-$}0.0249 & 12/12 & \makebox[1.3em][l]{$^{***}$} & \phantom{$-$}0.0270 & 10/12 & \makebox[1.3em][l]{$^{**}$} \\
    Hypernymy       & \phantom{$-$}0.0676 & 11/12 & \makebox[1.3em][l]{$^{**}$} & \phantom{$-$}0.0192 & 8/12 & \makebox[1.3em][l]{} & \phantom{$-$}0.0412 & 10/12 & \makebox[1.3em][l]{$^{**}$} & \phantom{$-$}0.0227 & 9/12 & \makebox[1.3em][l]{$^{*}$} \\
    Perceptual      & \phantom{$-$}0.0574 & 12/12 & \makebox[1.3em][l]{$^{***}$} & \phantom{$-$}0.0460 & 10/12 & \makebox[1.3em][l]{$^{**}$} & \phantom{$-$}0.0661 & 11/12 & \makebox[1.3em][l]{$^{**}$} & \phantom{$-$}0.0417 & 11/12 & \makebox[1.3em][l]{$^{**}$} \\
    Affective       & \phantom{$-$}0.0271 & 10/12 & \makebox[1.3em][l]{$^{*}$} & \phantom{$-$}0.0193 & 7/12 & \makebox[1.3em][l]{} & \phantom{$-$}0.0456 & 10/12 & \makebox[1.3em][l]{$^{**}$} & \phantom{$-$}0.0330 & 9/12 & \makebox[1.3em][l]{$^{*}$} \\
    \bottomrule
  \end{tabular*}
  \caption{\textbf{Basis privilege at different depths.} For each target and depth (normalized relative to the total number of layers), $\Delta r$ is the mean difference between native and random across models at the layer closest to the depth, and $n_{+}$ the number of models (out of 12) with a positive gap. Asterisks indicate statistical significance ($^{*}p<0.05$, $^{**}p<0.01$, $^{***}p<0.001$).}
  \label{tab:depth}
\end{table}

The model-to-model, model-to-feature and training dynamics analyses use a 60/20/20 split of the stimuli (train, validation, test), grouped by pair for the minimal pair tasks (Methods). The brain analyses use grouped 10-fold cross-validation over passages: Experiment 2 from \citet{pereira2018toward} holds out 9--10 of 96 passages (about 38 sentences), Experiment 3 holds out 7--8 of 72 passages (about 24 sentences); within each fold, we hold out 25\% of the training passages as the validation set. The scores we report are means over the 10 folds. 

In this paper we match on the absolute correlation. Matching on the signed correlation (as in \citealp{khosla2024privileged}) reproduces every conclusion of the paper except that the brain privilege in Experiment 3 becomes marginal. The difference comes from an artifact of GPT-2$_{XL}$, for which in late layers the representations are highly anisotropic: all units load with the same sign on a shared direction. With signed matching, a random rotation can be paired with voxels of either sign whereas the native basis cannot, so this asymmetry penalizes the native basis. After splitting, both $X$ and $Y$ are standardized by column (using the mean and standard deviation of the training split). For the random-rotation analyses we draw 10 independent $Q \sim \mathrm{Uniform}(\SOnn{})$ samples per condition and evaluate the matching score at 11 values of $\alpha \in \{0,\ 0.1,\ 0.2,\ \ldots,\ 1.0\}$, then average across the 10 random draws. The optimal axis alignment is trained with Adam, learning rate $10^{-4}$, one-cycle cosine schedule peaking at $\mathrm{lr}_{\max} = 10^{-3}$, 50 epochs, full-batch updates, with the orthogonality constraint enforced via the matrix-exponential parametrization described in the Methods. The best epoch is selected by mean matching score on the validation split, and we report the test-split matching score at that epoch.



\section{Layer-wise results}
\label{sec:si-layers}

The values we reported in the main text are averages across layers. Table~\ref{tab:depth} shows the basis privilege at four depths (for each model, the layer whose index divided by the number of layers is closest to 0.25, 0.5, 0.75 and 1). The gap is positive in 32 of 32 combinations of target $\times$ layer depth and significant in 28 combinations. The peak basis privilege varies by task, but it is significant across tasks at a depth of 1.0.

\section{BLiMP sub-tasks}
\label{sec:si-blimp}

\begin{figure}[t]
  \centering
  \includegraphics[width=\textwidth]{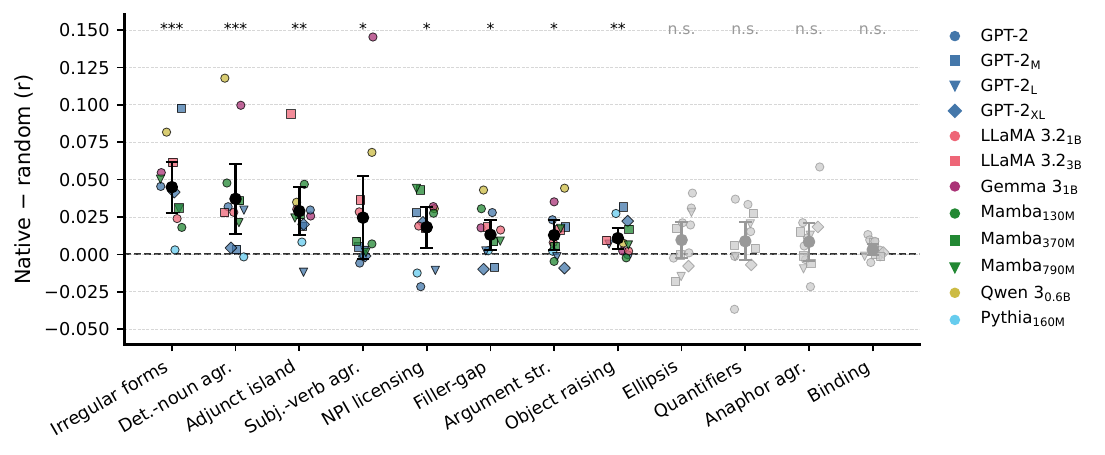}
  \caption{\textbf{Basis privilege for all BLiMP sub-tasks.} Difference between native and random alignment, averaged across layers, for each model. Symbols indicate the models; colors the model family; shapes the models within each family. Black dots and bars are the mean and 95\% confidence interval across models; asterisks mark statistical significance above zero ($^{*}p<0.05$, $^{**}p<0.01$, $^{***}p<0.001$).}
  \label{fig:S2}
\end{figure}

Fig.~\ref{fig:S2} shows the BLiMP result of Fig.~\ref{fig:2}B divided by the 12 sub-tasks (one for each broad phenomenon class; see Table~\ref{tab:blimp}). The gap between native and random, averaged across layers, is positive on average for every sub-task (range 0.003 to 0.045) and significantly greater than zero across models for eight sub-tasks. There appears to be bigger effects for contrasts involving morphosyntax (irregular forms, agreement) than for those that only involve structure (binding, raising, quantifiers).

\end{document}